\documentclass[letterpaper]{article} 
\usepackage[]{aaai2026}  
\usepackage{times}  
\usepackage{helvet}  
\usepackage{courier}  
\usepackage[hyphens]{url}  
\usepackage{graphicx} 
\usepackage{natbib}  
\usepackage{caption} 
\usepackage{algorithm}
\usepackage{algorithmic}
\usepackage{booktabs}
\usepackage{array}

\usepackage{newfloat}
\usepackage{listings}
\DeclareCaptionStyle{ruled}{labelfont=normalfont,labelsep=colon,strut=off} 
\floatstyle{ruled}
\newfloat{listing}{tb}{lst}{}
\floatname{listing}{Listing}
\title{Realist and Pluralist Conceptions of Intelligence \\and Their Implications on AI Research}
\author {
    Ninell Oldenburg \textsuperscript{\rm 1},
    Ruchira Dhar \textsuperscript{\rm 2},
    Anders Søgaard\textsuperscript{\rm 1,2}
}
\affiliations {
    \textsuperscript{\rm 1}Department of Philosophy, University of Copenhagen\\
    \textsuperscript{\rm 2}Department of Computer Science, University of Copenhagen\\
    niol@hum.ku.dk, rudh@di.ku.dk, soegaard@di.ku.dk
}

\usepackage{bibentry}

\begin{document}

\maketitle

\begin{abstract}
In this paper, we argue that current AI research operates on a spectrum between two different underlying conceptions of intelligence: \textit{Intelligence Realism}, which holds that intelligence represents a single, universal capacity measurable across all systems, and \textit{Intelligence Pluralism}, which views intelligence as diverse, context-dependent capacities that cannot be reduced to a single universal measure. Through an analysis of current debates in AI research, we demonstrate how the conceptions remain largely implicit yet fundamentally shape how empirical evidence gets interpreted across a wide range of areas. These underlying views generate fundamentally different research approaches across three areas. Methodologically, they produce different approaches to model selection, benchmark design, and experimental validation. Interpretively, they lead to contradictory readings of the same empirical phenomena, from capability emergence to system limitations. Regarding AI risk, they generate categorically different assessments: realists view superintelligence as the primary risk and search for unified alignment solutions, while pluralists see diverse threats across different domains requiring context-specific solutions. We argue that making explicit these underlying assumptions can contribute to a clearer understanding of disagreements in AI research.
\end{abstract}

\maketitle

\section{Introduction}

There is a growing debate among researchers on the nature and potential risks of AI, as well as the importance of respective research programs. On one hand, researchers warn that building bigger models may eventually lead to human-like, superhuman, or ``transformative'' and possibly dangerous levels of intelligence \citep{amodei2016concrete,bengio2025international,phan2025humanity,hendrycks2023overview}. On the other hand, scholars argue that simply creating larger models is unlikely to yield human-like cognition \citep{raji2021ai,mitchell2024debates,johnson2024imagining}.

What is intriguing about this debate is that researchers are observing the same empirical data, model architectures, scaling laws, and model performance, while holding on to profoundly different interpretations, e.g., scaling laws as a sign of emerging superintelligence vs. task-specific improvements. In this article, we suggest that this dichotomy can be traced back to diverging beliefs on what intelligence is. On one hand, intelligence realists hold that intelligence represents a single, universal capacity measurable across all systems. On the other hand, intelligence pluralists see intelligence as diverse, context-dependent capacities that cannot be reduced to a single universal measure. This has first-order implications for how we understand ``artificial'' intelligence, what we consider useful and dangerous, how we test, evaluate, and interpret systems, and how we build them. Different interpretations of empirical scaling laws, i.e., whether they represent convergence toward universal intelligence or domain-specific optimization, depend on which philosophical framework researchers inherently hold.

Our framework contributes to the philosophy of science by demonstrating how paradigmatic commitments shape observation and interpretation in contemporary AI research. Following \citet{kuhn1962nature}'s theory that theoretical frameworks determine what counts as significant evidence, we show how realist and pluralist assumptions function as competing paradigms that make identical empirical results (scaling curves, benchmark performance, system failures) intelligible in fundamentally different ways. This extends \citet{hanson1958logic}'s theory-ladenness of observation to AI: what researchers ``see'' in GPT-4's performance depends on their prior commitments about intelligence's nature. By making these implicit paradigms explicit, we enable more productive scientific discourse and clearer identification of genuine empirical disagreements vs. philosophical differences.

The remainder of this paper is structured as follows. We first present the features of each view with examples from the literature. We acknowledge that this conceptualization should be seen as a continuum rather than a hard classification before describing the implications these positions have on AI research methodology and AI alignment.

\begin{table*}[t]
\centering
\scriptsize
\small
\begin{tabular}{p{2.6cm}p{4.2cm}p{4.2cm}p{4.2cm}}
\toprule
\textbf{Dimension} & \textbf{Realist Indicators} & \textbf{Intermediate/Mixed} & \textbf{Pluralist Indicators} \\
\midrule
\textbf{Ontology} & Intelligence as singular capacity; seeks universal algorithms & Acknowledges diversity but looks for common principles & Intelligence as multiple, incommensurable capacities \\
\addlinespace
\textbf{Benchmarks} & Aggregates across domains (ARC-AGI, BIG-bench) & Domain-specific scores but with overall rankings & Separate evaluation per domain; refuses aggregation \\
\addlinespace
\textbf{Comparison Claims} & ``X is more intelligent than Y''; cross-species/-system rankings & Qualified comparisons (``more intelligent at task T'') & Rejects cross-system comparison as category error \\
\addlinespace
\textbf{Scaling Laws} & Evidence of AGI convergence & Sees progress but notes gaps & Benchmark-specific optimization \\
\addlinespace
\textbf{Failure Attribution} & ``Engineering problems''; ``needs more scale/data'' & Some architectural, some fundamental & ``Fundamental mismatches''; qualitative differences \\
\addlinespace
\textbf{AI Risk Focus} & AGI; univ. alignment principles & Both x-risk and near-term risks & Context-specific, near-term \\
\addlinespace
\textbf{Architecture Pref.} & Unified, general-purpose models & Modular systems & Specialized domain/task models \\
\addlinespace
\textbf{Success Criteria} & Domain-general competence & Task-specificity, some transfer & Specific cognitive phenomena \\
\bottomrule
\end{tabular}
\normalsize
\caption{Diagnostic Rubric for Identifying Intelligence Positions}
\label{tab:diagnostic-rubric}
\end{table*}

Our contribution is thereby not to invent a new distinction but to provide a \textit{targeted synthesis} that contextualizes long-standing debates in psychology, philosophy, and cognitive science within contemporary AI research. What is novel is demonstrating how these debates \textit{structure AI research in ways researchers often do not recognize}. By making explicit the implicit philosophical commitments underlying methodological choices, interpretive disagreements, and alignment strategies, we provide a vocabulary for clearer scientific discourse and more productive policy debates.


\section{Different Assumptions of Intelligence Assumptions}\label{sec:realismpluralism}

Before presenting our classification, we want to disclaim two points. First, by claiming that different researchers have different underlying assumptions of intelligence, we do not intend to imply that these necessarily have been \textit{articulated}. We posit that assumptions are revealed through \textit{methodological} choices and interpretation of empirical research, and will point below to these exact differences. Second, as mentioned above, by asserting that different views of intelligence exist, we do not intend to imply that these exist only under hard, declarative boundaries. Agreement with either position need not involve a wholesale endorsement of every aspect. We briefly develop an intermediate position below.


\subsection{Intelligence Realism}
Intelligence realism makes three distinct but related claims: (1) Ontological: There exists a single, universal computational process that constitutes intelligence across all systems; (2) Epistemological: This universal process is discoverable through scientific investigation; (3) Methodological: Different intelligent systems can be meaningfully ranked according to how well they approximate this universal process. Here, we detail its underlying assumptions.\footnote{We use ``Intelligence Realism'' to denote a position that combines ontological realism (intelligence exists as a real phenomenon) with monism (it constitutes a single, universal capacity). While these are technically separable commitments (some would say: orthogonal), with this notion, we posit that they travel together in the AI research context we analyze: Claims about universal intelligence algorithms typically presuppose their reality as computational processes.}

\paragraph{Algorithmic Universality}
The core ontological premise of intelligence realism is that intelligence follows universal and discoverable mathematical principles regardless of its instantiation. Epistemologically, this translates to the possibility of intelligence being described and tested for in abstract terms. One example of this perspective is Marcus Hutter's AIXI model \citep{hutter2003gentle}, which is a theoretical framework that defines (artificial general) intelligence as the ability to achieve goals in different environments. Hutter's AIXI formulation assumes that optimal intelligence can be expressed as a single optimization function across all environments, presupposing that different environments are commensurable and that a single policy exists that maximizes value across all contexts.

This implies that realists interpret scaling laws as evidence for algorithmic universality: if larger models consistently improve across diverse tasks, this suggests convergence toward a universal intelligence algorithm. The smooth power-law relationships observed across language modeling, reasoning, and multimodal tasks support the realist claim that a single optimization process underlies all intelligent behavior.

\paragraph{Single Optimality}
Closely tied in with the algorithmic universality assumption, secondly, intelligence realism ontologically asserts the existence of an \textit{optimal} intelligence algorithm. For this, consider intelligence as a universal algorithm that every instantiation (e.g., dolphin, human, or AI) can approximate.\footnote{``Instantiation'' refers to a particular physical or computational system that implements (more or less successfully) the abstract optimization function, much as different computers can instantiate the same sorting algorithm with varying efficiency.} If every instantiation of this algorithm is different, then every instantiation of this algorithm is closer or further away from \textit{the} universal intelligence algorithm. It follows that there exists a ranking of every organism of how closely this algorithm is instantiated. We want to note that single optimality does not posit that all systems can be linearly ranked, but that there exists an optimal mapping from computational resources to performance across environments. Different systems might be optimal under different resource constraints.

One way this position is methodologically reflected in the literature is again the AIXI work \citep{hutter2003gentle}. Here, expected rewards are counted over a specific time. The universal ``best'' algorithms have the highest number of expected rewards for a specific time span, while instantiations of a ``worse'' algorithm will count less. Also, Schmidhuber's work on an ``optimal ordered problem solver'' \citep{schmidhuber2004optimal}, a ``general and in a certain sense time-optimal way of solving one problem after another,'' suggests that there is an optimal way of solving problems and that this property can be discovered.

\paragraph{Intelligence Variance Through Implementation}
Following from the above, realism also asserts that different manifestations of intelligence are viewed as \textit{varying implementations} or approximations of the universal algorithmic core.\footnote{``Algorithmic core'' is the abstract computational procedure that is independent of physical substrate and defines the ``optimal'' information processing for a certain goal.} To be precise, realists do not assume identical algorithms, but rather that all intelligent behavior optimizes the same abstract objective function, as for example something like ``maximize expected utility given computational constraints'' \citep{legg2007universal,hutter2003gentle}. Different implementations (neural networks, biological brains) approximate this optimum differently, but the underlying optimization target is universal. This perspective is exemplified by comparative studies that attempt to measure intelligence across species using information-theoretic or computational metrics.

Consider the work of \citet{commons2008toward}, which developed cross-species intelligence metrics based on task-solving efficiency and adaptive behavior. Their research suggests that despite significant differences in neural architecture, different species might be understood as instantiating similar underlying computational principles. A chimpanzee solving a tool-use problem, a corvid caching food, and an AI system playing chess could potentially be analyzed through a common algorithmic lens.

\paragraph{Reducibility}
Any advanced form of intelligence can be reduced to a combination of constitutive features. By doing so, it could reveal fundamental principles applicable across all intelligent systems. This reductionist approach finds methodological expression in efforts to create comprehensive intelligence benchmarks across different specific domains.

For example, \citet{goertzel2014artificial} presents an attempt to develop tests that measure intelligence independent of a specific implementation. The most comprehensive effort in this direction is \citet{hernandez2017measure}, which develops formal frameworks for evaluating diverse intelligent systems (from biological organisms to AI) using task-based measures that claim to transcend specific implementations. Similarly, \citet{chollet2019measure} proposes the Abstraction and Reasoning Corpus (ARC) as a measure of general fluid intelligence focused on skill-acquisition efficiency rather than task-specific performance. By designing tasks that require flexible problem-solving across diverse domains, researchers sought to create a ``universal intelligence test'' that could compare cognitive capabilities across different systems that transcend domain-specific knowledge.

\paragraph{Commensurability}
The final key assumption following the assumptions on variation through implementation and reducibility is that all forms of intelligence can, after all, be meaningfully compared along a single dimension or a small set of unified dimensions. This perspective underpins many intelligence tests, both natural and artificial. Without commensurability, the realist's claim that there exists an optimal intelligence algorithm becomes meaningless: if different forms of intelligence cannot be compared, then the notion of optimality loses its foundation.

As an example, \citet{legg2007universal} argue that not only artificial intelligence but also intelligence across different species can be formalized as ``an agent’s ability to achieve goals in a wide range of environments.'' More specifically, their formal definition of universal intelligence $\Upsilon$ of an agent $\pi$, $\Upsilon(\pi) = \sum_{\mu\in E} w_\mu V_\mu^\pi $ encodes the realist assumption of commensurability by assigning weights $w_\mu$ to different environments $\mu$ of the space of all environments $E$. This implies that these environments can be meaningfully compared and aggregated into a single intelligence measure.\footnote{Historically, in human intelligence, this was also expressed as the g-factor \citep{eid2017anomalous}, or the IQ \citep{terman1916measurement}. The debate over whether g represents a real cognitive capacity or merely a useful statistical artifact mirrors the realism-pluralism distinction we develop here. Realists treat g as reflecting a unified cognitive capacity \citep{jensen1998factor}, while pluralists see it as an instrumental construct that may not correspond to any natural kind (see also \citet{borsboom2005measuring} and \citet{vessonen2019representing}).}


\subsection{Intelligence Pluralism}
Intelligence pluralism denies all three realist claims, arguing instead that: (1) Ontologically, intelligence consists of multiple, incommensurable computational processes rather than varying implementations of a single process; (2) Epistemologically, these processes can only be understood \textit{within} their specific ecological and evolutionary contexts because context is constitutive, not merely obscuring; (3) Methodologically, cross-system comparisons of ``intelligence'' are therefore either impossible or meaningless because there is no universal standard against which to measure.\footnote{These pluralist commitments connect to broader traditions in cognitive science. Gardner's theory of multiple intelligences \citep{gardner1983frames}, Gigerenzer's work on ecological rationality \citep{gigerenzer1996reasoning}, and enactivist approaches emphasizing embodied, embedded cognition \citep{wilson2002six} all challenge the notion of intelligence as a unified, context-independent capacity. These frameworks differ in details, yet share the core pluralist insight that intelligent behavior emerges from specific agent-environment couplings rather than universal computational principles.}

\paragraph{Algorithmic Diversity}
Ontologically, pluralism sees intelligence as inherently tied to the specific context, environment, and evolutionary history in which it develops. While realism assumes that there is one universal algorithm across species that intelligence can be represented as, pluralism assumes that there are many different algorithms across species and maybe even within species. 

One such example is the navigational system of desert ants. Desert ants have evolved a path integration system that is adapted to their featureless environment and calculates their return path using measurements of distance and direction traveled. This strategy would be suboptimal in more complex terrains where agents can rely on landmark-based navigation \citep{wehner2003path}. This challenges the realist assumption of optimality. The ant's path integration system would be catastrophically suboptimal in forest environments where landmark navigation dominates due to potential obstacles. Yet both systems are ``optimal'' within their contexts. This poses a dilemma for realists: either (1) there's no universal optimum, or (2) optimality is context-dependent, which undermines the universality claim.

Realists might posit that pluralists confuse implementation diversity with algorithmic diversity: Yes, ants and humans navigate differently, but both implement approximate solutions to the same abstract problem: optimal path planning under uncertainty. The diversity is in implementation, not in the underlying computational problem. 

Pluralists reject this move as an ontological error. The realist position assumes that ``path planning'' exists as a natural kind, i.e., as a problem discoverable through formal analysis that exists independently of any particular agent (see \citet{magnus2012scientific}; \citet{boyle2024disagreement}). Pluralists see this as an analyst's construction retrospectively imposed on diverse behaviors. Ants are not solving ``path planning'' but following chemical gradients that happened to be evolutionarily successful; the ``problem'' only exists from an external perspective that imposes human-like goals onto non-human systems. For pluralists, path-finding in ants and humans does not yield different solutions to the same problem. Rather, they solve fundamentally different problems because they are different agents in different environments. Methodologically, this means we cannot identify the ``same'' problem across systems because the problem itself is constituted differently for each agent-environment coupling.

\paragraph{Multiple Equilibria}
Rather than a single optimal algorithm, pluralism posits that numerous valid cognitive architectures exist, optimized for different niches and challenges. For instance, different bird species have evolved distinct spatial memory systems: food-caching corvids like Clark's nutcrackers develop extraordinary spatial memory for storing thousands of seed locations, while other birds rely on entirely different navigational strategies \citep{sherry1992spatial}. In AI, this translates to recognition that different machine learning architectures might be optimally suited to different problem domains. A neural network excelling at image recognition may be fundamentally ill-suited to causal reasoning tasks, and so on.

This aligns with bounded rationality approaches that emphasize how cognitive architectures are optimized for specific computational constraints and environmental regularities rather than universal optimality \citep{simon1990bounded,lieder2020resource}. Different resource constraints (time, memory, energy) lead to qualitatively different optimal strategies, undermining the notion of a single best algorithm.

It further implies that these different strategies, as we will detail below, cannot be ranked. As a first intuition, consider the navigational skills of migratory birds, the social skills of elephant herds, or the distributed intelligence of slime molds, of which each represents a cognitive strategy that defies ranking or comparison. We will also detail below what implications this has on the meaningfulness of the word ``intelligence''.

\paragraph{Emergence}
Pluralism posits that intelligence emerges from dynamic interactions between agents and environments, rather than from a single set of abstract principles that can be \textit{universally} generalized in mathematical form.\footnote{Pluralists do not deny that intelligent behavior can be modeled mathematically: ant navigation and human planning can both be formalized. Rather, they deny that these diverse formalizations can be reduced to or unified under a single universal mathematical framework.} It assumes that every form of behavior is a way of solving some problem for a specific type of agent in a specific type of environment. In contrast to realism, it posits that even abstract optimization targets embed specific assumptions about what counts as ``utility'' and which computational constraints matter. These assumptions are inevitably shaped by particular evolutionary histories and environmental pressures, making supposedly universal targets actually context-dependent.

Realists would posit that emergence does not preclude universal principles. Physics has universal laws despite emergent phenomena. Why should intelligence not have universal computational principles despite contextual emergence? Pluralists would counter that physical universals govern simple interactions that scale up to complexity. ``Intelligence'', however, may be ontologically different: intentional states, semantic content, and goal-directed behavior may be irreducibly tied to \textit{specific forms} of embodiment and environmental embedding (see \citet{penny1995embodied,cangelosi2015embodied}), not merely difficult to study in the abstract but constituted by these specificities. The analogy to physics assumes precisely what's in dispute: that intelligence involves universal building blocks.

Here, the obvious question is whether every species that has emerged over millions of years can be counted as intelligent. If intelligence means ``cognitive strategies that enable successful environmental adaptation,'' then any species that has survived evolutionary pressures possesses intelligence. If pluralism rejects universal standards for intelligence and instead evaluates systems within their own contexts, then most surviving biological systems meet the criteria for intelligence within their respective niches.

It follows that under pluralism, the term ``intelligence'' loses discriminatory power. If every successful cognitive strategy counts as ``intelligent,'' then intelligence becomes equivalent to ``any cognitive adaptation that works''. A concept that applies to everything explains nothing and becomes analytically vacuous. Can this then still be a helpful classification to look at, e.g., risks of AI under the pluralist lens? The implication is that \textit{other} properties become more analytically useful, such as danger for other agents, efficiency given the environment and agent features, and innovativeness in comparison with other agents. Importantly, these criteria are all \textit{relational} in the sense that they are not universal or general but only valid for a specific set of species. We expand more on (emergent) LLM capabilities below under \textit{Implications for AI Research: Interpretation}.

\paragraph{Irreducibility}
Pluralism further posits that intelligence as a whole cannot be reduced to abstract concepts but has to be viewed in its environmental and agent-based complexity. For example, some principles of human intelligence provide a scaffolding for understanding some principles of AI, but the complexity of the whole ``intelligence'' may be fundamentally irreducible to mathematical principles: only task-by-task can an algorithm be extracted from a human problem-solving attempt and then emulated in an artificial system.

Famously, octopuses demonstrate a distributed intelligence where a significant portion of their neural processing occurs in their arms, which is a very different cognitive architecture from centralized brain-based intelligence \citep{godfrey2016other}. This challenges realist reducibility assumptions more fundamentally. If intelligence can be ``distributed'' across arms with semi-autonomous processing, what is the unit of analysis? The realist must either (1) deny that arm-based processing constitutes intelligence or (2) accept that intelligence does not reduce to centralized algorithms, which undermines the universal algorithm thesis. Ontologically, this suggests intelligence is not a property of algorithms per se but of specific agent-environment-body configurations. Methodologically, this means we cannot extract ``the intelligence algorithm'' from the octopus any more than we can extract ``the wetness algorithm'' from water; it's an emergent property of the whole system that cannot be isolated from its constituent relations. 

A realist might counter that this misidentifies the relevant unit of analysis: the unit is the optimization process, not the physical substrate, and that distributed processing just shows that optimal intelligence can be physically distributed while remaining computationally unified. However, pluralists might see this, again, as an ontological error: the realist move to abstract away from physical substrates assumes the very thing in question: that intelligence exists independently of its material and environmental instantiation. When the premise is to compare different systems, the specific agent-environment-body configuration is not an optional detail to be abstracted away but constitutive of what intelligence is in each case.

An example from the computational sciences is the field of computational cognitive science that methodologically tries to model the functions of the human brain \textit{one algorithm at a time}. Examples are understanding epistemic language \citep{ying2025understanding}, virtual bargaining \citep{levine2024rules}, logical reasoning \citep{olausson2023linc}, online goal inference \citep{zhi2020online}, or knowledge inference in lie production \citep{tan2024reasoning}. In contrast to realist methodologies, these works are trying to figure out the exact algorithm for this one specific problem rather than trying to infer proxies for ``intelligence'' for a range of tasks from one model.

\paragraph{Incommensurability}
The last key assumption is that pluralism assumes that different forms of intelligence may be fundamentally ``alien'' to each other, prohibiting straightforward comparisons and understanding along universal dimensions. This follows from what we established above: if every system implements a range of specific, environmentally optimized solutions to its specific problem, a comparison to other systems would have to take into account this exact agent-environment pair. However, as soon as we compare it to the exact pair, we compare it to itself. As elaborated above, this also means that different systems cannot be ranked according to how ``intelligent'' they are and, with a high likelihood, can also not understand each other's ``intelligence'' as a whole, if this terminology is still useful at all. While realists would argue for an ``optimization under constraints'', pluralists would posit that this still assumes environments and constraints are commensurable enough to define meaningful optima. But if cognitive strategies are genuinely incommensurable, the entire optimization framework breaks down.

As a small example, researchers took a very long time until octopuses were ascribed something like ``intelligent'' behavior, and very likely this is still under an anthropomorphic lens. As an implication for neural language models, this suggests that these systems might develop forms of ``intelligence'' qualitatively different from human cognition. The models that excel at text generation, for instance, operate through fundamentally different mechanisms than human linguistic cognition. In this famous view, the term ``understanding'' would be inherently misplaced and overgeneralized when talking about a feature for artificial systems as they perform statistical transformations while lacking core aspects of human reasoning (see further \citep{mitchell2021ai}.


\subsection{Intermediate Positions}

The realism-pluralism distinction operates across multiple dimensions (ontological, epistemological, methodological), and researchers need not adopt consistent positions across all dimensions. For instance, one might hold \textit{ontological realism} (believing intelligence reflects a universal computational process) while practicing \textit{methodological pluralism} (using domain-specific evaluation frameworks because universal metrics are currently infeasible). Conversely, one might accept \textit{ontological pluralism} (multiple incommensurable forms of intelligence) while employing \textit{methodological reductionism} (attempting to decompose each form into constituent algorithms for practical analysis).

\section{Implications for AI Research}

Having outlined the realism-pluralism spectrum, we will now turn to the implications that different inherent positions on intelligence have on research methodology, the interpretation of results, and safety and governance. 

\subsection{Methodology}

Methodological implications are the most direct and immediate consequences of an underlying intelligence position. Here, we look at model selection, benchmark design, and validation and success criteria.

\paragraph{Model Selection}

Even though both views agree on there being one brain in humans, the exact way in which this property can be realized in artificial models is the core of the realist-pluralist debate. Realist assumptions favor unified architectures capable of general-purpose large language models, universal approximators, or architectures that can be scaled across diverse tasks. The popularity of transformer architectures partly reflects their apparent domain-generality.

Pluralist assumptions favor specialized architectures optimized for specific cognitive domains. Pluralists do not seek universal solutions but develop targeted approaches (as exemplified above): understanding epistemic language \citep{ying2025understanding}, virtual bargaining \citep{levine2024rules}, logical reasoning \citep{olausson2023linc}, online goal inference \citep{zhi2020online}, or knowledge inference in lie production \citep{tan2024reasoning}. Emulating a whole brain would be putting together all these specialized architectures in an efficient way \citep{griffiths2024bayesian}.

\paragraph{Benchmark Design}

Realist assumptions naturally lead toward unified, domain-general benchmarks that assume different cognitive tasks tap into a common underlying capacity. BIG-Bench \citep{kazemi2025big} or ARC-AGI \citep{chollet2019measure} exemplify this approach. They assume that diverse domains across law, medicine, mathematics, or social reasoning (for BIG-Bench) or, again ``intelligence'' as a whole (for ARC-AGI) can be meaningfully aggregated to a single measure and thereby implicitly assume the domains share sufficient commonality to warrant aggregated scoring. However, multi-task benchmarks also show the spectrum nature of our framework. They do acknowledge domain distinctions (pluralist view) and yet still aggregate performance into unified rankings that assume commensurable underlying capacities (realist view), which reflects the intermediate positions we have touched upon above.

Pluralist methodology, by contrast, demands task-specific evaluations that respect the ecological context in which ``intelligence'' arises and operates. Rather than seeking a universal metric, pluralists insist on separate evaluation frameworks for each cognitive domain, designed around the specific environmental pressures and adaptive challenges that shaped those capacities. Examples of this specialization assumption are cognitive benchmarks like ConceptARC for abstract concept understanding \citep{moskvichev2023conceptarc}, FANToM for Theory of Mind, \citep{kim2023fantom}, NormAd for adaptive norm understanding \citep{rao2024normad}, or EWoK for world knowledge \citep{ivanova2024elements}.

\paragraph{Validation and Success Criteria}

Realists judge success in terms of progress toward universal principles, i.e., theories that explain intelligence across species, domain-general competence, or across environments. Famous examples include AIXI \citep{hutter2003gentle}, \citet{hernandez2010measuring}, or \citet{legg2007universal}. They would further validate AI capabilities by demonstrating equivalence or superiority to biological intelligence on standardized tasks, as they assume commensurability between systems. Here, of course, the level of abstraction and the differentiation in architecture and functionality are of utmost importance. Some intermediate positions would argue that while architectural or fine-grained comparisons might be less meaningful, the system can functionally be compared or at some higher level of abstraction, as we detailed above.

Pluralist methodology judges success in terms of understanding \textit{specific} cognitive phenomena within their ecological contexts. Rather than seeking universal explanations, pluralists aim for a rich understanding of how particular cognitive strategies solve specific adaptive challenges. For artificial systems, pluralists argue that these cannot be validated against biological intelligence because they solve fundamentally different computational problems through categorically different mechanisms. This, of course, creates big challenges for safety and governance: if we cannot assess how a machine is doing on metrics that make sense to us, how can we assess whether it will become dangerous or not? We will detail this below in the section on safety and governance.

\subsection{Interpretation}

Beyond shaping research methodology, realist and pluralist assumptions fundamentally alter how empirical evidence gets interpreted. The same data patterns become evidence for opposing theoretical positions, exemplifying what \citet{kuhn1962nature} called paradigm-dependent observation. This is not merely an abstract philosophical point. In practice, realist and pluralist researchers examine the same systems, read the same papers, and observe the same benchmark results, yet reach opposite conclusions about what these findings mean for AI capabilities and risks. The framework we developed provides a vocabulary for identifying when disagreements are empirical (solvable by more data) versus paradigmatic (requiring explicit philosophical examination).

\paragraph{Scaling Laws} 

Two examples of opposite interpretations of the same data, scaling laws, are the realist ``Sparks of Artificial General Intelligence'' \citep{bubeck2023sparks} and the pluralist ``Why AI is Harder Than We Think'' \citep{mitchell2021ai}. 

\citet{bubeck2023sparks}'s central claim is that GPT-4 ``could reasonably be viewed as an early (yet still incomplete) version of an artificial general intelligence (AGI) system,'' which rests on the premise that intelligence constitutes a unified capacity manifesting across diverse cognitive domains. Multi-domain competence is interpreted as evidence for an underlying general intelligence progressing along a measurable continuum toward full AGI. The aggregated score reflects genuine cross-domain reasoning ability that approximates general intelligence. 

The smooth scaling curves that show consistent improvement with increased parameters and training data are read as confirmation of algorithmic universality. Sudden capability jumps at critical scales are interpreted as genuine cognitive emergence, where quantitative parameter increases trigger qualitative leaps in reasoning ability \citep{wei2022emergent}.

\citet{mitchell2021ai} critiques these claims by proposing how some, as she claims, narrow advances in task-mastery are seen as the ``first step'' towards some more general goal, even though this progress might not lie on a continuum and an unexpected obstacle might always be in the way (\citet{mitchell2021ai} citing \citet{dreyfus2012history}). For example, Deep Blue or GPT-2 have been claimed to be the first steps towards an ``AI Revolution'' \citep{aron2016ai} or ``general intelligence'' \citep{alexander2019step}. She supports this with an analogy by the engineer and Hubert Dreyfus' brother, Stuart Dreyfus: ``It was like claiming that the first monkey that climbed a tree was making progress towards landing on the moon'' \citep{dreyfus2012history}.

Under pluralist interpretation, scaling behaviors reflect sophisticated statistical pattern matching reaching various thresholds rather than genuine cognitive emergence. Multi-domain competence indicates that large datasets contain sufficient statistical regularities for pattern matching across diverse text types, not that the system possesses domain-general reasoning capabilities (see \citet{kambhampati2025reasoning,kambhampati2025stop}). The ``emergence'' of new capabilities represents accumulated statistical sophistication crossing perceptual thresholds, not the development of genuine understanding \citep{schaeffer2023emergent}. Lastly, domain-specific performance ceilings reflect the inherent limitations of statistical approaches to cognitive tasks requiring genuine understanding (e.g., \citet{isbilen2022statistical}).

\paragraph{Emergent Capabilities} 

This interpretive difference extends to the very concept of ``emergent abilities'' in LLMs, which has become a focal point of the realism-pluralism debate. What realists interpret as discontinuous capability emergence, i.e., the sudden appearance of capabilities like arithmetic, theory of mind, or complex reasoning at scale, pluralists challenge as an artifact of measurement assumptions. First, apparent discontinuities may result from human-designed benchmarks with arbitrary pass/fail thresholds that make smooth capability development appear sudden \citep{schaeffer2023emergent}. The ``emergence'' lives in the metrics, not the underlying process. 

Second, and more fundamentally, pluralists question whether LLMs and humans develop ``the same'' capabilities at all. When an LLM produces arithmetic outputs, is it ``doing arithmetic'' in any sense comparable to human calculation, or performing statistical pattern matching that we retrospectively label as ``arithmetic'' because outputs coincide? The realist assumes ``arithmetic'' is a natural computational category that both humans and LLMs instantiate; the pluralist sees this as imposing human categories onto alien processes. This connects to benchmark selection: we evaluate LLMs on human-centric tasks and interpret their success as evidence of universal intelligence, assuming our capabilities represent the right targets for measuring intelligence. Realists interpret scaling laws as evidence of convergence toward universal intelligence; pluralists see them as evidence that we've successfully optimized for our own benchmarks.

\paragraph{Failure Analysis}
Realists frame observed deficits as architectural flaws or insufficient size, both of which can be solved by engineering and what some have termed ``hardware fixes'' \citep{musser2018job}. When GPT-4 exhibits limitations in long-term planning, temporal reasoning, or causal understanding, \citet{bubeck2023sparks} characterize these as ``missing components'' amenable to engineering solutions through ``further training.'' This treats general intelligence as achievable through scaling solutions rather than inherent ecological and qualitative advances.

Pluralists view such failures as revealing qualitative differences between human and artificial cognition that resist technical remediation rather than being incidental. In particular, they reveal fundamental architectural mismatches between statistical pattern matching and the contextual, embodied cognition that characterizes biological intelligence. For instance, they argue that large language models still struggle with common-sense physical reasoning \citep{ivanova2024elements}, exhibit systematic failures in Theory of Mind \citep{kim2023fantom}, or do vision processing in an inherently different, non-human way \citep{bowers2023deep}. They argue that systematic failures indicate fundamental limits, suggesting that human-like cognition requires qualitatively different architectures rather than incremental improvements to existing systems (e.g., \citet{diaz2024scaling}; \citet{griffiths2024bayesian})

\subsection{Alignment}

These interpretive differences have direct consequences for AI safety and governance.

\paragraph{Risk Assessment}
Realist assumptions naturally lead to concerns about superintelligent agents pursuing unified goals with domain-general capabilities. If intelligence constitutes a universal algorithm that can be optimized across all environments, and we see that some models are improving on one axis, then sufficiently advanced AI systems will surpass humans on that general intelligence scale. Systems will necessarily develop coherent goal structures and general competencies that enable them to pursue objectives across diverse domains \citep{bostrom2014superintelligence}. Sufficiently advanced AI systems would develop universal instrumental goals, making control increasingly difficult as capabilities scale \citep{amodei2016concrete}.

This realist framework directly informs major alignment research programs. For instance, the Constitutional AI framework assumes harmful behaviors emerge from a single, trainable value system that can be shaped through universal principles \citep{bai2022constitutional}. Similarly, scalable oversight presupposes that alignment properties transfer across capability levels, i.e., if we can align a weaker system, we can use it to align stronger systems because they implement the same underlying optimization process \citep{zeng2025redefining,bowman2022measuring}.

The realist assumption of algorithmic universality also underlies concerns about mesa-optimization and deceptive alignment \citep{hubinger2019risks}. If intelligence follows universal principles, then sufficiently capable systems will inevitably develop internal optimization processes that may not align with their training objectives, leading to systematic deception during training that only manifests during deployment.

Pluralist assumptions generate fundamentally different threat models focused on diverse system behaviors emerging from the interaction of specialized AI systems operating in different contexts. The core premise that pluralists argue for is that the currently used psychometric tests investigate the wrong tasks while failing to capture crucial aspects of intelligence, such as meta-learning, causal reasoning, innate priors, and robust generalization, all considered essential for superintelligence \citep{raji2021ai,mitchell2021ai}. This leads to different risk priorities. Since current AI systems already demonstrate ``intelligence'', i.e., high sophistication, \textit{within} specific domains like algorithmic bias in criminal justice, labor displacement \citep{gebru2024tescreal}, or misinformation \citep{bender2024resisting}, immediate societal impacts in these domains might take precedence over \textit{hypothetical} superintelligence scenarios \citep{moorosi2023ai}.

However, as touched upon above, the boundaries are not clear-cut, and there might be many researchers and research programs that acknowledge both superintelligence risk and societal risk (e.g., see \citet{hendrycks2023overview}).

\paragraph{Alignment Approaches}

Realists often seek universal principles to control or align systems and favor solutions like reward models that capture human preferences across all domains \citep{russell2019human,zhi2024beyond}, interpretability techniques that reveal the universal optimization process underlying AI behavior \citep{elhage2022toy}, and robustness guarantees that hold regardless of deployment context \citep{zhao2024improving,sanfiz2021benchmarking}. Some of these attempts, however, face challenges that suggest the validity of a more pluralist view, including issues of context dependence \citep{milliere2023alignment}, frame problems \citep{shanahan2004frame,peterson2025context}, and specification challenges \citep{zhi2024beyond}. For example, impact measures like Attainable Utility Preservation \citep{turner2020conservative} struggle to define impacts without reference to specific contexts and value systems. The persistent challenge of reward hacking across different alignment approaches further demonstrates how safety mechanisms designed under universalist assumptions often fail when confronted with the rich complexity of real-world environments \citep{amodei2016concrete}.

Pluralist priorities emphasize context-specific alignment: developing evaluation frameworks that assess AI behavior within specific ecological niches \citep{enevoldsen2023danish,martin2019camembert}, specific cognitive functions \citep{zhi2024pragmatic,olausson2023linc}, and specific human-AI collaboration contexts rather than general-purpose intelligence \citep{collins2024building,zhi2024pragmatic,oldenburg2024learning}. This includes research on cultural value diversity in AI systems \citep{rao2024normad,li2024culturellm,li2024culturepark}, work on developing domain-specific safety measures for AI deployment in healthcare \citep{aggarwal2023artificial}, autonomous driving \citep{waschle2022review}, education \citep{clark2025building}, etc., and investigation of how different AI architectures might be optimally suited to different human social contexts \citep{tomavsev2020ai,abbass2019social}.

\paragraph{Governance Implications}

These different approaches lead to concrete policy disagreements. Realists tend to support capability-based regulation focusing on general AI capabilities regardless of application, and international coordination on universal AI safety standards \citep{hendrycks2023overview}. This perspective informs proposals for global AI governance institutions and capability-based licensing schemes.

Pluralists advocate for application-specific regulation that varies by domain and context, emphasizing democratic participation in AI governance and regulatory frameworks that account for the diversity of human values and social contexts \citep{bogiatzis2024beyond}. The EU AI Act exemplifies this approach by categorizing risks based on deployment context rather than general capability level \citep{veale2021demystifying}.


\section{Conclusion}

We have argued that contemporary AI research operates along a spectrum between two fundamentally different conceptions of intelligence that remain largely implicit: Intelligence Realism and Intelligence Pluralism. We showed how they fundamentally shape the field's methodology, interpretation, and risk assessment. Making these underlying assumptions explicit serves several purposes.

\textbf{Methodologically}, recognizing the realism-pluralism spectrum helps researchers understand why different approaches to benchmark design, model selection, and validation criteria persist. These are not merely engineering choices but reflect deep commitments about the nature and existence of ``intelligence''. Researchers favoring a domain-specific architecture and task-specific, cognitive benchmarks probably agree more with the pluralist framing than those favoring general models and aggregate benchmarks. More importantly, they are operating from different ontological starting points.

\textbf{Interpretively}, the framework clarifies why identical empirical evidence generates opposite conclusions. When realists and pluralists examine GPT-4's performance, they are not disagreeing about the data but about what ``intelligence'' is and how it can be recognized. This explains why scaling law debates, failure analysis, and capability assessments remain unresolved despite abundant empirical work. The disagreements are paradigmatic, not merely empirical.

\textbf{For AI risk research}, these different conceptions generate fundamentally different risk models and governance approaches. Realists focus on superintelligence scenarios and universal alignment principles; pluralists emphasize distributed, context-specific, and currently existing risks and tailored safety measures. Both perspectives identify risks, but they prioritize different threats and solutions. Making these commitments explicit enables more productive debates: rather than arguing past each other about ``the'' AI risk, stakeholders can identify which conception of intelligence underlies their concerns and whether disagreements are empirical or philosophical.

We do not claim to resolve the realism-pluralism debate that would require extensive empirical and philosophical work beyond this paper's scope. Rather, our contribution is to make explicit the implicit assumptions that structure contemporary AI research, providing a vocabulary for clearer scientific and policy discourse. As AI systems become more capable and consequential, understanding these foundational disagreements becomes increasingly urgent. A research community that explicitly engages with its philosophical commitments is better positioned to navigate the challenges ahead.

\section*{Acknowledgements}
We thank members and visitors of the Hong Kong AI \& Humanity lab for their fruitful comments and feedback on earlier versions of this paper, especially Rachel Sterken, Seth Lazar, Nate Sharadin, Herman Cappelen, and Kate Vredenburgh. We also thank the five anonymous reviewers for their constructive feedback that helped to shape the paper.

\bibliography{refs}

\end{document}